# Early Bearing Fault Diagnosis of Rotating Machinery by 1D Self-Organized Operational Neural Networks

Turker Ince[1], Junaid Malik[2], Ozer Can Devecioglu[2], Serkan Kiranyaz[3], *Senior Member, IEEE*, Onur Avci[4], Levent Eren[1] and Moncef Gabbouj[2], *Fellow, IEEE.*

[1]Electrical & Electronics Engineering Department, Izmir University of Economics, Izmir, Turkey
[2]Department of Computing Science, Tampere University, Tampere, Finland
[3]Department of Electrical Engineering, Qatar University, 2713 Doha, Qatar
[4]Department of Civil, Construction and Environmental Engineering, Iowa State University, Ames, USA

Corresponding author: Turker Ince (e-mail: turker.ince@ieu.edu.tr).

**ABSTRACT** Preventive maintenance of modern electric rotating machinery (RM) is critical for ensuring reliable operation, preventing unpredicted breakdowns and avoiding costly repairs. Recently many studies investigated machine learning monitoring methods especially based on Deep Learning networks focusing mostly on detecting bearing faults; however, none of them addressed bearing fault severity classification for early fault diagnosis with high enough accuracy. 1D Convolutional Neural Networks (CNNs) have indeed achieved good performance for detecting RM bearing faults from raw vibration and current signals but did not classify fault severity. Furthermore, recent studies have demonstrated the limitation in terms of learning capability of conventional CNNs attributed to the basic underlying linear neuron model. Recently, Operational Neural Networks (ONNs) were proposed to enhance the learning capability of CNN by introducing non-linear neuron models and further heterogeneity in the network configuration. In this study, we propose 1D Self-organized ONNs (Self-ONNs) with generative neurons for bearing fault severity classification and providing continuous condition monitoring. Experimental results over the benchmark NSF/IMS bearing vibration dataset using both x- and y-axis vibration signals for inner race and rolling element faults demonstrate that the proposed 1D Self-ONNs achieve significant performance gap against the state-of-the-art (1D CNNs) with similar computational complexity.

## I. INTRODUCTION

Electrical rotating elements and machines are widely used in various industrial and commercial applications on account of their reliability and efficiency. Mechanical bearing faults have the highest statistical occurrence percentage among all the motor fault types. Effective condition monitoring and early fault detection and diagnosis of RM is critical for maintaining reliable operation, avoiding unpredicted breakdowns, reducing operating costs and improving productivity. Generally, the main approaches to fault detection and diagnosis (FDD) can be classified as analytical model-based methods, signal-based methods and knowledge-based methods: Model-based methods develop mathematical models with parameters based on physical working principles of the complex systems and/or measured data through system identification and state-space modeling [1]. In practice, for complex machine systems, it is increasingly difficult to model its input-output behavior and harder to estimate its parameters [2]. The signal-based methods are based on typical signal analysis such as vibration, motor current, speed, and temperature, using signal processing methods such as fast Fourier transform [3], spectral estimation [4], time-frequency [5] and wavelet transformation [6], sequence analysis [7] and scale-invariant feature transform (SIFT) [8]. Motor current signature analysis (MCSA) is an example of a successful monitoring technique for electrical machines based on acquisition and analysis of electrical signatures for motor currents [4]. The signal-based FDD methods, similar to the model-based FDD, require *a priori* knowledge of signal patterns and often advanced signal processing tools with increased computational complexity need to be applied effective fault diagnosis [9]. The data-driven or knowledge-based systems, as opposed to the previous two methods, utilize large volumes of collected data (but only small amount having label information) enabled by the advances in data acquisition and control systems without the need for explicit models [2]. Although many data-driven techniques including shallow machine learning methods proposed in the literature reported satisfactory levels of fault detection and diagnosis performance, they depend on hand-designed features and/or classifiers of different types and usually used small amount of motor or RM data. Thus, they cannot accomplish generic solution as their performance inevitably degrades for different types of systems or faults and for larger datasets.

Recently, modern data-driven deep learning models developed by the machine learning researchers have been proposed as solution to FDD problems. As opposed to shallow supervised learners, deep learning networks (DNNs) can learn the required features from the raw input data via training fully automatically voiding the need for the handcrafted statistical or transform-domain feature representations [10]-[17]. However, for a proper training, they need well-labelled training datasets with massive size. Jia et al. [10] showcased a five-layer DNN model for fully automatic intelligent FDD of RM. The pretraining of DNN model is achieved by an unsupervised autoencoder (AE) and the vibration signal frequency spectra was utilized as input. On the benchmark Case Western Reserve University bearing dataset, their method obtained 95.8% classification accuracy to differentiate three bearings with ball defects under different loads. In [11], a deep large memory storage retrieval (LAMSTAR) neural network which processes short-term Fourier transform (STFT) of raw acoustic emission signals for bearing fault diagnosis is presented. From the test results using the real data, the LAMSTAR network method improves performance compared to CNNs at both the normal and relatively low input shaft speeds. Xia et al. [12] proposed an intelligent FDD method by a DNN based on stacked denoising. Fault diagnosis testing accuracy of larger than 97% was achieved on the CWRU dataset and their method can still obtain over 95% diagnosis accuracy with only small fraction (3%) of labelled data. However, they employed a complex five-layer DNN with input layer of 600 neurons and three hidden layers of 400, 200 and 50 neurons, respectively. For gearbox fault identification, Chen et al. [13] developed a deep CNN with different manually selected sets of statistical and spectral (FFT) features at its input and achieved classification accuracies ranging from 89% to 98% for various sets of inputs. In [14], the three-layer CNN model processed scaled discrete Fourier transform (DFT) of raw accelerometer signals to obtain 87.25% average accuracy for bearing fault detection. More recently, A hierarchical adaptive deep CNN (ADCNN) based FDD approach was proposed by Guo et al. [15] and achieved 99% classification accuracy over the CWRU dataset. However, this approach has slow convergence and high computational complexity. Generative adversarial networks (GANs)-based framework was proposed by Shao et al. [16] to learn to generate 1D realistic raw data from mechanical sensor signals to augment real sensor data to resolve the issues of unbalanced data on training deep networks for applications in machine fault diagnosis. In [17], an improved convolutional deep belief network (CDBN) was proposed for diagnosis of rolling bearing faults where the original vibration signal was first transformed to the frequency-domain via FFT before being input to an optimized model structure to achieve better accuracy than traditional SAE, ANN, DBN and CBDN models. Such deep models [15]-[17] have a high computational complexity in common which prevents their usage in low-power computing environments in real-time. Additionally, they have not dealt with small training rate (small train data size compared to test data) which is usually the case for practical applications.

In order to incorporate the aforementioned issues, in this study, we draw the focus on the bearing fault severity classification where the objective is to maximize early fault diagnosis performance when the data is scarce and the network complexity is minimized for a real-time application over any platform. In our previous study in [18], we proposed for the first time, a compact 1D CNN for real-time motor bearing fault classification and achieved the *state-of-the-art* performance demonstrated over the benchmark bearing fault dataset using raw current signals. Shallow 1D CNNs have light training process with only few dozens of back-propagation (BP) epochs, performing the classification task easily in real-time [18]. Hence, they are good fit for real-time RM condition monitoring and advance FDD systems.

Recent studies [19]-[21] have pointed out that CNNs having homogenous network configuration based on a first-order neuron model cannot adequately learn problems with a complex and highly nonlinear solution space [19]-[21] unless a sufficiently high network depth and complexity (variants of CNN) are accommodated. Recently, Self-Organized Operational Neural Networks (Self-ONN) have been proposed to achieve a high heterogeneity level with self-organized operator optimization capability to maximize the learning performance [25]. The superior regression capability of Self ONNs over image segmentation, restoration, and denoising was demonstrated in recent studies [25], [29].

In this study, to achieve an elegant computational efficiency and network heterogeneity, we propose one-dimensional Self-organized ONNs (1D Self-ONNs) with the generative neuron model for RM bearing fault severity classification. Having low computational complexity is important since most FDD algorithms are implemented on multi-function and/or multi-device monitoring systems. Our objective is to improve RM bearing fault classification performance compared to compact 1D CNNs [18] while preserving real-time computational ability. In summary, the contributions of the paper are the following:
- 1D Self-ONN model with enhanced learning capacity is first-time proposed for early bearing fault diagnosis in RMs.
- The run-to-failure bearing vibration signals from the benchmark NSF/IMS dataset are utilized to detect early-level, moderate-level and severe-level faults with higher accuracy using 1D Self-ONNs.
- The proposed 1D Self-ONN classifier has a compact structure (with fewer neurons and parameters) with low computational complexity which enables its use in more complex multi-function and/or multi-device monitoring systems.
- This study shows that improved RM bearing fault classification performance can still be achieved without application of complex deep networks needing dropout and data augmentation techniques (due to data scarcity).

The rest of the paper is organized as follows: Section II presents novel 1D Self-ONNs by comparing to the standard 1D CNNs and ONNs. The methodology of the proposed RM bearing fault severity classification and condition monitoring system based on 1D Self-ONN model is presented in Section III. In Section IV, a detailed set of experimental results based on the benchmark motor vibration dataset are discussed and the performance of the proposed technique is assessed against the state-of-the-art approach with 1D CNNs. Finally, in Section V, the conclusions are presented and topics for future research are listed.

## II.1D SELF-ORGANIZED OPERATIONAL NEURAL NETWORKS

In this section, we will proceed by revisiting how ONNs generalize the 1D convolution operation. Afterwards, the mathematical model of the proposed generative neuron-based 1D Self-ONN will be presented. First, we consider the case of the $k^{th}$ neuron in the $l^{th}$ layer of a 1D convolutional neural network. To be concise, we assume the 'same' convolution operation with unit stride and the required amount of zero padding. The output of this neuron can be formulized as follows:

$$x_k^l = b_k^l + \sum_{i=0}^{N_{l-1}} x_{ik}^l \qquad (1)$$

where $b_k^l$ is the bias associated with the this neuron and $x_{ik}^l$ is defined as:

$$x_{ik}^l = Conv1D(w_{ik}, y_i^{l-1}) \qquad (2)$$

Here, $w_{ik} \in \mathbb{R}^K$ is the kernel connecting the $i^{th}$ neuron of $(l-1)^{th}$ layer to $k^{th}$ neuron of $l^{th}$ layer, while $x_{ik}^l \in \mathbb{R}^M$ is the input map, and $y_i^{l-1} \in \mathbb{R}^M$ are the $l^{th}$ and $(l-1)^{th}$ layers' $k^{th}$ and $i^{th}$ neurons' outputs respectively. By definition, the convolution operation of (2) can be expressed as follows:

$$x_{ik}^l(m) = \sum_{r=0}^{K-1} w_{ik}^l(r) y_i^{l-1}(m+r) \qquad (3)$$

where $w_{ik}$ kernel and shifted versions of $(l-1)^{th}$ layer's $i^{th}$ neuron's output $y_i^{l-1}$ are element-wise multiplied and summed for the length of the kernel to obtain values of $M$-dimensional input vector $x_{ik}^l$.

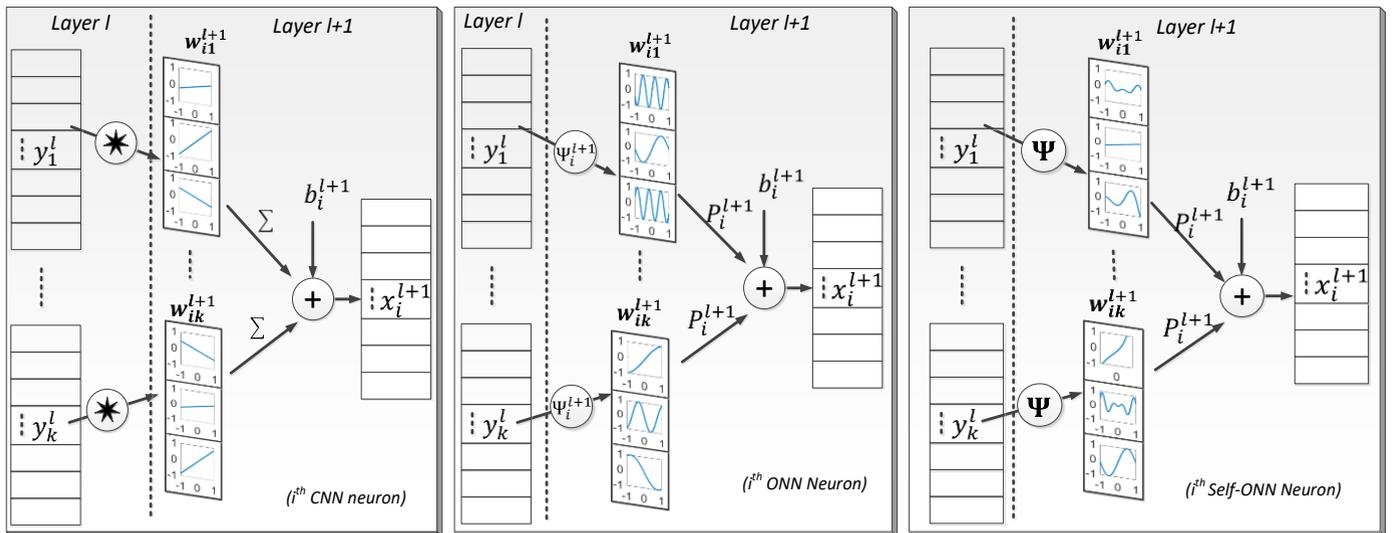

Figure 1: Depiction of the 1D nodal operations with the 1D kernels of the kth CNN (left), ONN (middle) and Self-ONN (right)

The main idea behind an operational neuron is a generalization of the CNNs' convolutional neuron above as follows:

$$\overline{x_{ik}^l}(m) = P_k^l \left( \psi_k^l \left( w_{ik}^l(r), y_i^{l-1}(m+r) \right) \right)_{r=0}^{K-1} \quad (4)$$

where $\psi_l^k(\cdot): \mathbb{R}^{M \times K} \to \mathbb{R}^K$ and $P_k^l(\cdot): \mathbb{R}^K \to \mathbb{R}^1$ are termed as nodal and pool functions, respectively, assigned to the $k^{th}$ neuron of $l^{th}$ layer. Note that $r$ represents the running index for the $K \times 1$ kernel vector $w_{ik}^l$, which is operated by the nodal operator $\psi_l^k(\cdot)$ optimized for the $k^{th}$ neuron of the $l^{th}$ layer during the Backpropagation training.

In a heterogenous ONN configuration, every neuron has uniquely assigned $\psi$ and $P$ operators. Owing to this, an ONN network has the advantage of having the flexibility of incorporating any non-linear transformation suitable for the given learning problem at hand. However, hand-crafting a suitable library of possible operators and searching for an optimal one for each neuron in a network introduces a significant overhead, which rises exponentially with increasing network complexity. Moreover, it is also possible that the right operator for the given learning problem cannot be expressed in terms of well-known functions. The self-organization ability is inherently embedded in the generative neuron model and thus, a Self-ONN will have the nodal operator functions optimized during the training procedures to maximize the learning performance as illustrated in Figure 1 illustrates convolutional and operational neurons of a CNN and ONN with fixed (static) nodal operators, while the generative neuron can have any arbitrary nodal function, $\Psi$, (including possibly standard types such as linear and harmonic functions) for each kernel element of each connection.

Operational neuron comes as a non-linear extension and generalization of the CNNs' convolutional neuron which performs solely the linear convolution operation. The 1-D Self-ONN aims to have the potential to achieve superior operational diversity and flexibility through optimized nodal and pool operators during the Backpropagation training and hence maximize its learning performance. In this study, we propose a 1D Self-ONN model for the challenging problem of RM bearing fault severity classification from raw vibration signals thanks to its superior learning ability over CNNs.

To formulate a nodal transformation which does not require a pre-selection and manual assignment of operators, we make use of the Taylor-series based function approximation, which is given for an infinitely differentiable function $f(x)$ about a point $a$ as:

$$f(x) = \sum_{n=0}^{\infty} \frac{f^{(n)}(a)}{n!}(x-a)^n \quad (5)$$

The $Q^{th}$ order truncated approximation of (5), formally known as the Taylor polynomial, takes the following form:

$$f(x)^{(Q,a)} = \sum_{n=0}^{Q} \frac{f^{(n)}(a)}{n!} x^n \quad (6)$$

The above formulation enables the approximation of any function $f(x)$ sufficiently well in the close vicinity of $a$. If the coefficients $\frac{f^{(n)}}{n!}$ are tuned and the inputs are bounded, the formulation of (6) can be used to generate any transformation. This is the key idea behind the *generative neurons* which form Self-ONNs. Specifically, in terms of notation used in (4), the nodal transformation of a generative neuron would take the following general form:

$$\widetilde{\psi_k^l}\left(w_{ik}^{l(Q)}(r), y_i^{l-1}(m+r)\right) = \sum_{q=1}^{Q} w_{ik}^{l(Q)}(r,q) \left(y_i^{l-1}(m+r)\right)^q \quad (7)$$

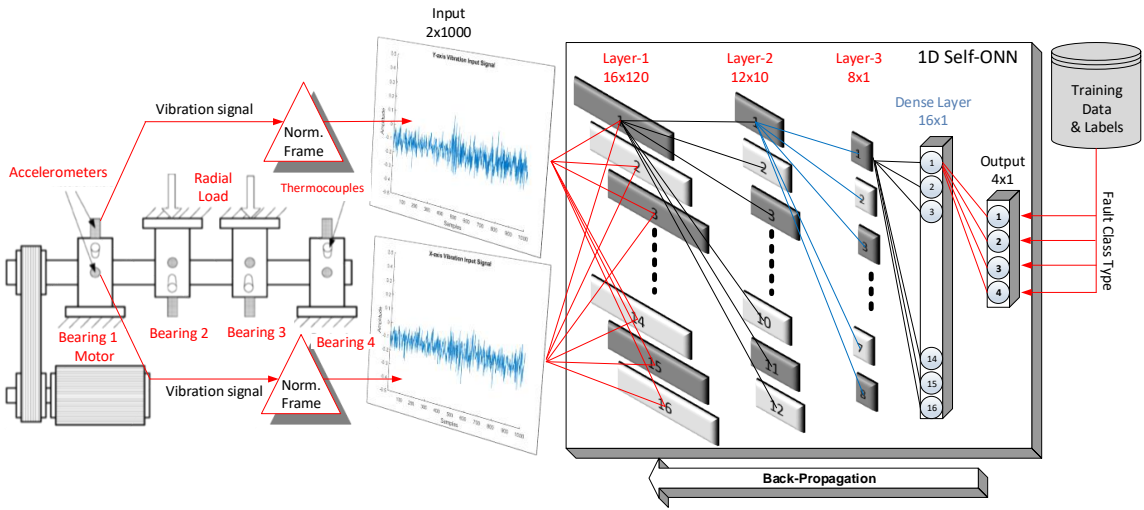

**Figure 2: Experimental setup per the IMS Bearing Dataset [26] and the general view of the proposed methodology with training (offline) and online bearing fault level detection steps.**

In (7), $Q$ is a hyperparameter which controls the degree of the Taylor approximation, and $w_{ik}^{l(Q)}$ is a learnable kernel of the network. A key difference in (7) as compared to the convolutional (3) and operational (4) model is that $\widetilde{\psi_k^l}$ is not fixed, rather a distinct operator over each individual output, $y_i^{l-1}$, and thus requires $Q$ times more parameters. Therefore, the $K \times 1$ kernel vector $w_{ik}^l$ has been replaced by a $K \times Q$ matrix $w_{ik}^{l(Q)} \in \mathbb{R}^{K \times Q}$ which is formed by replacing each element $w_{ik}^l(r)$ with a $Q$-dimensional vector $w_{ik}^{l(Q)}(r) = [w_{ik}^{l(Q)}(r,0), w_{ik}^{l(Q)}(r,1), \dots, w_{ik}^{l(Q)}(Q-1)]$. The input map of the generative neuron, $\tilde{x}_{ik}^l$ can now be expressed as,

$$\widetilde{x_{ik}^l}(m) = P_k^l \left( \sum_{q=1}^{Q} w_{ik}^{l(Q)}(r,q) \left(y_i^{l-1}(m+r)\right)^q \right)_{r=0}^{K-1} \quad (8)$$

During training, as $w_{ik}^{l(Q)}$ is iteratively tuned by back-propagation (BP), *customized* nodal transformation functions will be generated as a result of (8), which would be tailored for $i-k^{th}$ connection. This enables an enhanced flexibility which provides three key benefits. First, the need for manually defining a list of suitable nodal operators and searching for the optimal operator for each neuronal connection is naturally alleviated. Secondly, the heterogeneity is not limited to each neuronal connection $i-k$ but down to each kernel element as $\tilde{\psi}_l^k\left(w_{ik}^{l(Q)}(r), y_i^{l-1}(m+r)\right)$ will be unique $\forall r \in [0,1,\dots,K-1]$. Note that such a diversity is not achievable even with the flexible operational neuron model of ONNs. Thirdly, in generative neurons, the heterogeneity is driven only by the values of the weights $w_{ik}^{l(Q)}$ and the core operations (multiplication, summation) are the same for all neurons in a layer, as shown in (8). Owing to this, unlike ONNs, the generative neurons inside a Self-ONN layer can be parallelized more efficiently, leading to a considerable reduction in computational complexity and time. The generalized formulations of the forward-propagation through a Self-ONN neuron and back-propagation training of the Self-ONNs are described in [25], [29].

## III. METHODOLOGY

The mechanical bearing faults and electrical failures per insulation or winding faults are among the main sources of faults in induction machines (IMs) and/or RMs. Bearing faults have the highest frequency of occurrence and are one of the most challenging to detect and diagnose. As mechanical defects, bearing faults result in vibrations at certain characteristic fault frequencies, which can be obtained from the shaft speed and the bearing geometry [27]. Interestingly, despite being the most challenging to detect and quantify, bearing defects are the least expensive to fix if detected in time to be replaced [6]. Therefore, in this study, we focus on the accurate and early detection and hence severity classification of RM bearing faults.

The IMS bearing dataset [26] has been frequently used as a benchmark dataset to test and validate diagnostic algorithms. As shown in Figure 2, the test setup for the IMS dataset has four bearings installed on a shaft. In this study, we used the first dataset for which data was collected from two accelerometers attached to the bearing housing. The experiment was carried out until failure of the bearings that worked for more than hundred million revolutions occur. As such, failures occurred as a result of accumulation of incremental damages on the inner race, outer race and roller element as a whole, with the test-to-failure loadings. The effects of the faults are also visible on the amplitude spectrum of the vibration signals. The formulations for characteristic vibration frequencies along with bearing faults spectrum are presented in Supplementary.

The general view of the proposed motor bearing fault severity classification approach is shown in Figure 2 where 1D Self-ONNs receives the normalized vibration frames of 1000 samples from the two accelerometers on each bearing. The normalized raw vibration signals from two accelerometers (orthogonal x- and y- directions of the bearings) are partitioned into the non-overlapping frames of 1000 samples. Each frame is first standard normalized to take out the effect of the DC offset and amplitude bias, then scaled linearly in the range of [-1,1] to form the input frames of the Self-ONN classifier. For the recorded vibration signals used for training and evaluation, the four classes indicating the status of the motor are: *healthy, early-level fault, moderate-level fault* and *severe-level fault*. Here, there are three frequency zones analyzed in the bearing vibration spectrum to assess the condition of a bearing. The shaft velocity related frequencies, bearing defect frequencies, and bearing natural resonance frequencies appear in zones I, II, and III, respectively.

While the *healthy* bearing will have energy content only in the zone I, the *early-level fault* is associated with the energy in zones I and III. For *moderate-level fault*, the bearing fault characteristic vibration frequencies become present in Zone II and the energy levels in zone III increase. Finally, the *severe-level fault* is associated with increased energy levels in zones I and III; and bearing fault characteristic fundamental and harmonic vibration frequencies appear to be more pronounced in zone II [30]. In our study, experts labeled the vibrations signals for three different fault severity levels in advance based on the spectral analysis. Further details about the bearing fault characteristic vibration frequencies are provided in the Supplementary Material section.

Maintenance Systems (IMS) [26]. The test rig setup includes four double row bearings attached on a shaft with a constant rotational velocity of 2000 RPM. A radial load of 6 kips was applied onto the shaft and bearing by a spring mechanism. PCB accelerometers (353B33 high sensitivity Quartz ICP) were attached on the bearing housing. For dataset 1, two accelerometers were attached for each bearing's x- and y-axes; for datasets 2 and 3, one accelerometer was attached

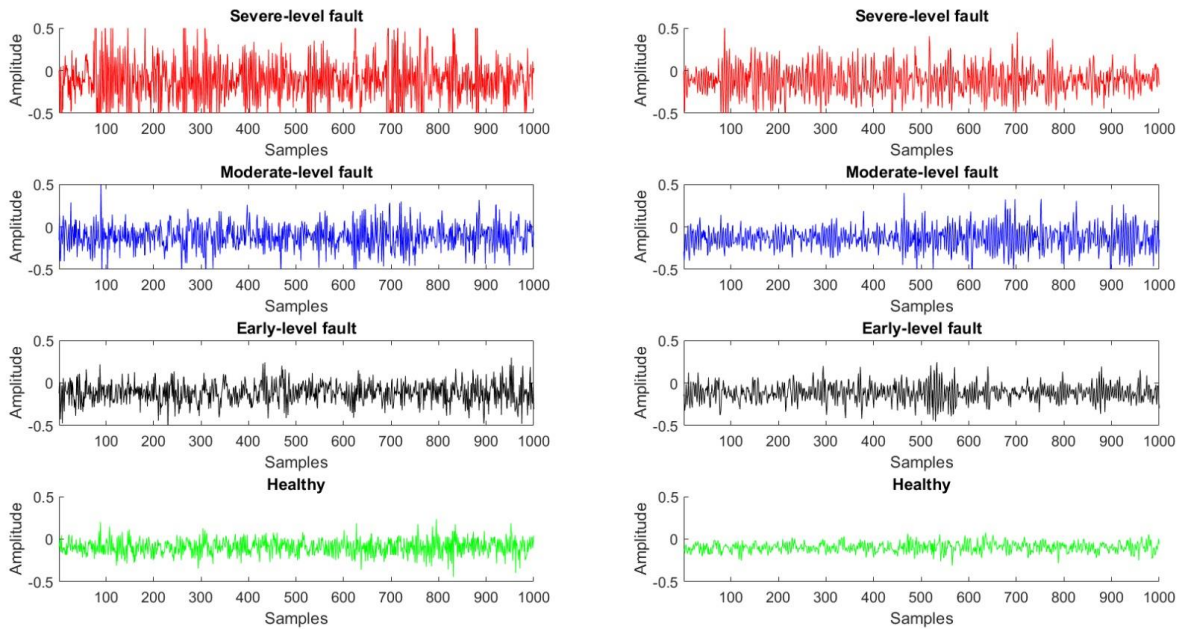

Figure 3: Typical frames of vibration signals for different levels of inner race (left) and rolling element faults (right).

As stated earlier, the 1D CNN is a special case of 1D Self-ONN with $Q = 1$ assigned for all neurons. Therefore, the experimental setup shown in Figure 2 can also conveniently be used to evaluate the performance of a conventional 1D CNN. The experimental setup and network parameters will be presented in the next section.

## IV. EXPERIMENTAL RESULTS

In this section, the benchmark motor bearing fault dataset used in this study will be introduced first. Then, the experimental setup used for evaluation of the proposed bearing fault severity classification framework will be presented. The comparative evaluations and the overall results of the experiments obtained using the real motor vibration signals will be presented as a following step. Additionally, the computational complexity of the proposed approach for both training (offline) and classification (online) phases will be evaluated in detail and compared with the 1D CNNs.

### A. The Bearing Fault Dataset

The benchmark bearing dataset used in this study was generated by the NSF IUCRC, Center for Intelligent for each bearing. Each recorded file (NI DAQ Card 6062E) is made up of 20,480 points with a sampling frequency of 20.48 kHz and it describes a test-to-failure experiment. One second data acquisition was made every ten minutes, except for the first dataset for which the first forty-three files were acquired every five minutes. Generally, three frequency components, the shaft speed related frequencies, the bearing defect frequencies, and the bearing natural resonance frequencies, appear in the spectrum of the bearing vibration signals according to the condition of a bearing. For the case of a bearing defect, the bearing fault fundamental frequencies appear and as the severity of the fault increases the energy levels of the shaft speed related and the harmonic frequencies also increase. Additionally, these frequency components can be stronger or weaker depending on the type of fault (i.e. inner race vs. rolling element) which can make bearing fault diagnosis more challenging.

For experimental evaluation, we used the whole 36-minute-long dataset 1 from the test-to-failure experiment which contains inner race and roller element damages which occurred in bearings 3 and 4, respectively (there are two accelerometers for each bearing). After the test-rig runs for a while, some deterioration in healthy bearings starts to occur and becomes more pronounced with increasing fault severity over time. Towards the end of the endurance test, some defect types become evident in these bearings. Sample time-domain vibration signals for different levels of inner race and rolling element faults are shown in Figure 3.

From the analysis of vibration signals, the following initiation of the early level faults' impulsive signals and more chaotic (noise-like) vibrations with higher amplitudes can be clearly seen as the level of the fault increases. As opposed to earlier studies that applies decimation (hence reducing the original sampling rate) and filtering before normalization in the preprocessing stage, in this study, the frames of raw vibration signal at its original sampling rate (20.48 kHz) is collected from x- and y-axes of motor's bearings. The nonlinear activation function "*tanh*" is used in all MLP neurons. The kernel size and the subsampling factor for the first layer of Self ONN are set as 41 and 8, for the second layer as 41 and 8, and for the third layer as 9 and 2, respectively. The max-pooling is used in all sub-sampling layers. We implemented the proposed 1D Self ONN classifier using FastONN library [28] which utilizes Python and PyTorch. For training, ten-fold cross-validation method is used to enhance generalization and prevent over-fitting. For all experiments, the maximum number of BP iterations (epochs) was set to 50. Early-stopping based on minimum rain classification error threshold of 3% is used in order to avoid over-fitting. Using the SGD Optimizer, the learning

**Table I: Inner race fault severity classification performance of 1D Self-ONNs**

| Q | Configuration | Healthy | | | Early-level fault | | | Moderate-level fault | | | Severe-level fault | | |
|---|---|---|---|---|---|---|---|---|---|---|---|---|---|
| | | *Sen* | *Ppr* | *F1* | *Sen* | *Ppr* | *F1* | *Sen* | *Ppr* | *F1* | *Sen* | *Ppr* | *F1* |
| 1 | (16-12-8)+(16-4) | **1.0000** | **1.0000** | **1.0000** | 0.9406 | 0.9500 | 0.9453 | 0.8284 | 0.8325 | 0.8304 | 0.8909 | 0.8775 | 0.8841 |
| 1* | (32-24-16)+(16-4) | **1.0000** | **1.0000** | **1.0000** | 0.9509 | 0.9675 | 0.9591 | 0.8921 | 0.8475 | 0.8692 | 0.9007 | 0.9300 | 0.9151 |
| 3 | (16-12-8)+(16-4) | **1.0000** | **1.0000** | **1.0000** | 0.9750 | 0.9750 | 0.9750 | 0.9282 | 0.8725 | 0.8995 | 0.9033 | **0.9575** | 0.9296 |
| 5 | (16-12-8)+(16-4) | **1.0000** | 0.9975 | 0.9987 | 0.9775 | **0.9775** | 0.9775 | **0.9295** | 0.8900 | 0.9093 | 0.9139 | 0.9550 | 0.9340 |
| 7 | (16-12-8)+(16-4) | **1.0000** | **1.0000** | **1.0000** | **0.9848** | 0.9725 | **0.9786** | 0.9146 | **0.9100** | **0.9123** | 0.9263 | 0.9425 | **0.9343** |
| 9 | (16-12-8)+(16-4) | **1.0000** | **1.0000** | **1.0000** | 0.9823 | 0.9700 | 0.9761 | 0.9116 | 0.9025 | 0.9070 | 0.9218 | 0.9425 | 0.9320 |

**Table II: Rolling element fault severity classification performance of 1D Self-ONNs**

| Q | Configuration | Healthy | | | Early-level fault | | | Moderate-level fault | | | Severe-level fault | | |
|---|---|---|---|---|---|---|---|---|---|---|---|---|---|
| | | *Sen* | *Ppr* | *F1* | *Sen* | *Ppr* | *F1* | *Sen* | *Ppr* | *F1* | *Sen* | *Ppr* | *F1* |
| 1 | (16-12-8)+(16-4) | 0.9925 | 0.9975 | 0.9950 | 0.9607 | 0.9775 | 0.9690 | 0.8966 | 0.8450 | 0.8700 | 0.8865 | 0.9175 | 0.9017 |
| 1* | (32-24-16)+(16-4) | **1.0000** | **1.0000** | **1.0000** | **0.9899** | 0.9850 | **0.9875** | 0.9139 | 0.9025 | 0.9082 | 0.9140 | 0.9300 | 0.9219 |
| 3 | (16-12-8)+(16-4) | **1.0000** | **1.0000** | **1.0000** | 0.9778 | **0.9900** | 0.9839 | **0.9435** | 0.8775 | 0.9093 | 0.9054 | **0.9575** | 0.9307 |
| 5 | (16-12-8)+(16-4) | **1.0000** | 0.9975 | 0.9987 | 0.9706 | **0.9900** | 0.9802 | 0.9005 | 0.8825 | 0.8914 | 0.9102 | 0.9125 | 0.9114 |
| 7 | (16-12-8)+(16-4) | **1.0000** | **1.0000** | **1.0000** | 0.9752 | 0.9850 | 0.9801 | 0.9306 | 0.9050 | **0.9176** | 0.9312 | 0.9475 | **0.9393** |
| 9 | (16-12-8)+(16-4) | **1.0000** | **1.0000** | **1.0000** | 0.9800 | 0.9800 | 0.9800 | 0.9071 | **0.9275** | 0.9172 | **0.9463** | 0.9250 | 0.9355 |

normalized and used as input for the classifiers.

### B. Experimental Setup

The compact 1D Self ONN with only 3 operational layers and 2 dense (MLP) layers is used in the proposed bearing fault severity classification framework (Figure 2) for all experiments. Therefore, the proposed solution is computationally efficient and well-suited for real-time FDD applications. The 1D Self ONN configuration has 16, 12, and 8 neurons in the 3 hidden operational layers, respectively, and 6 neurons in the hidden MLP layer. Meanwhile, the MLP output layer size is fixed as 4 which is the number of classes. The two-neuron input layer takes the input vibration signals factor, $\varepsilon$, is set to 0.2. Mean-Squared-Error (MSE) is used for BP error calculation. For each data partition, we repeat five individual BP runs and the average fault classification results are reported.

### C. Comparative Evaluations

For evaluation, the following three commonly used performance metrics in the literature are used: Precision or Positive Predictivity (*Ppr*), Recall or Sensitivity (*Sen*) and F1-score (*F1*). These metrics are distinctive for each class and they assess the capability of the proposed classifier to distinguish specific events from non-events. *Precision* is the rate of correctly classified events among all detected events; *Recall* is the rate of correctly classified events in all events, and *F1-score* is the harmonic mean of the model's Precision and Recall. The formulations for these performance metrics in terms of false negatives (FN), false positives (FP), true negatives (TN) and true positives (TP) can be expressed as follows:

$$Ppr = \frac{TP}{TP+FP}, \quad Sen = \frac{TP}{TP+FN}, \quad (9)$$

$$F1 = \frac{2 \times Ppr \times Sen}{Ppr + Sen}$$

Moreover, comparative evaluations with the state-of-the-art 1D CNN method are performed in [18]. Recall that 1D CNN is equivalent to 1D Self-ONN with Q=1. Specifically, from the dataset 1, 20 files for each of four classes (healthy, early-level fault, moderate-level fault and severe-level fault) are appropriately selected by the experts through spectral analysis. In this study, for each class, a total of 400 frames of 1000 samples are used for the evaluation (training and testing) of the proposed model. For each fold in the 10-fold cross-validation, a total of 360 frames per class were used for training, while the remaining data was used testing. The inner-race and rolling-element fault severity level classification performance results of the proposed framework using the standard metrics are presented in Table I and Table II. For inner-race fault classification, Self-ONNs have a significant superiority in all metrics resulting around 3-8% F1 gain over the 1D CNN with the same configuration. Performance gain is more significant for moderate and severe-level fault recognition. In this case, 1D Self-ONN classifier with Q=7 achieved the best classification performance. For rolling element faults, Self-ONNs have slightly lower (2-4%) F1 performance gains over the CNN. The 1D Self-ONN classifier with Q=7 also gives the best performance for this case. When the number of convolutional neurons are doubled, the performance of the CNN (indicated as 1* in Table I and Table II) has improved slightly; however, its recognition performance is still worse than the Self-ONN with half the neurons especially for moderate to severe-level inner-race faults. This further shows the superiority of the 1D Self-ONNs over the CNNs for this problem.

Table III presents sample confusion matrices per fault case corresponding to the best performing 1D Self-ONNs and CNNs. Since both moderate-level and severe-level bearing faults have energy content in zones II and III, it is more difficult to separate the transition point from moderate to severe level in a run-to-failure setup. The superiority of Self-ONNs over the CNNs is also visible especially in differentiating medium- and severe-level faults.

**Table III: Sample confusion matrices of 1D Self-ONNs and 1D CNNs (in parenthesis) for two fault types.**

**Inner Race Fault (Q=7)**

| | | Classification Result | | | |
|---|---|---|---|---|---|
| | | H | ELF | MLF | SLF |
| Ground Truth | H | 400 (400) | 0 (0) | 0 (0) | 0 (0) |
| | ELF | 0 (0) | 389 (380) | 11 (20) | 0 (0) |
| | MLF | 0 (0) | 6 (24) | 364 (333) | 30 (43) |
| | SLF | 0 (0) | 0 (0) | 23 (49) | 377 (351) |

**Rolling Element Fault (Q=7)**

| | | Classification Result | | | |
|---|---|---|---|---|---|
| | | H | ELF | MLF | SLF |
| Ground Truth | H | 400 (399) | 0 (1) | 0 (0) | 0 (0) |
| | ELF | 0 (3) | 394 (391) | 6 (6) | 0 (0) |
| | MLF | 0 (0) | 10 (15) | 362 (338) | 28 (47) |
| | SLF | 0 (0) | 0 (0) | 21 (33) | 379 (367) |

### D. Computational Complexity Analysis

For computational complexity analysis of the proposed classifier model, we provide the formulation for calculating the total number of multiply-accumulate operations (MACs) and the total number of parameters (PARs) of a generative neuron inside a 1D Self-ONN. To calculate the number of trainable parameters, we recall from Section II that, for each kernel connection, the generative neuron has $Q$ times more learnable parameters than the 1D CNN. Cumulatively, the number of trainable parameters, $n_k^l$, of the $k^{th}$ neuron of $l^{th}$ layer, is as follows:

$$n_k^l = N_{l-1} * K_k^l * Q_k^l \quad (10)$$

In (10), $N_{l-1}$ is the number of neurons in layer $l-1$, $K_k^l$ is the kernel size used in the neuron and $Q_k^l$ is the approximation order selected for this neuron. Finally, to calculate the total number of MAC operations, in order to produce a single element in the output, we require $K_k^l * Q_k^l$ MAC operations for each output map of the previous layer [29].

Generalizing this, we can write the following:

$$MAC_k^l = N_{l-1} * \left| \widetilde{x_{ik}^l} \right| * K_k^l * Q_k^l \quad (11)$$

where |·| is the cardinality operator. For notational convenience, the bias term and the cost of Hadamard exponentiation are omitted from (11). All the experiments were carried out on a 2.2GHz Intel Core i7-8750H with 8 GB of RAM and NVIDIA GeForce GTX 1050Ti graphic card. Both training and testing phases of the classifier were processed using the GPU. Along with the average time complexity, using the formulations in (10) and (11), we provide the overall PARs and MACs for both network models in Table **IV**.

**Table IV: Network models and average classification times.**

| Network | Layer 1 size | Layer 2 size | Layer 3 size | PARs | MACs (M) | Avg. Time (μs) |
|---|---|---|---|---|---|---|
| 1D CNN* | 32 | 24 | 16 | 37980 | 5.078 | 5.4 |
| 1D CNN | 16 | 12 | 8 | 10296 | 1.908 | 5 |
| 1D Self-ONN (Q=7) | 16 | 12 | 8 | 70584 | 13.253 | 9 |

For the GPU implementation, specifically, the total time for the classification (FP) of a 50ms (1000 samples) input segment is 9 μsec for a Self-ONN classifier with Q=7 (for both inner-race and rolling element faults detection). Such a computation speed (> 5000 times of real-time speed) naturally allows a real-time execution even on low-power mobile devices. As a secondary important advantage of the proposed 1D Self-ONN, this real-time performance attributed to the compact nature of the network along with its high learning capacity may enable the diagnosis of more complex systems made of several RM components with the same tool.

## V. CONCLUSIONS

In this study, a new RM bearing fault severity classification and condition monitoring system is proposed based on a novel 1D Self-ONN model. The *self-organizing* capability of the proposed network voids the need for prior operator search runs and presents a crucial computational advantage over ONNs. Our objective is to push the frontier set by the landmark study [18] based on adaptive 1D CNNs by achieving *the state-of-the-art* bearing fault classification performance with an elegant computational efficiency. Each generative neuron of a Self-ONN can individually optimize the nodal operator function of each kernel element resulting in a neuron-level heterogeneity maximizing the network diversity and performance. Overall, in this new-generation machine learning paradigm, the traditional weight optimization in conventional CNNs has entirely been transformed into an operator optimization process. Nevertheless, as demonstrated in this study, Self-ONNs can still be implemented using 1D convolutions and, a Self-ONN and a CNN with the same configuration have a similar computational complexity with respect to the theoretical parallelizability of the operations.

The proposed system is tested with real bearing vibration signals from the well-known IMS bearing dataset and the experimental results reveal its efficacy and potential for providing continuous condition monitoring and bearing fault severity classification. Low computational complexity of the proposed FDD algorithm also leaves more bandwidth for other possible metering/monitoring functions implemented on the system. In a fair evaluation with comparable network configurations, Self-ONNs outperform CNNs with a significant margin in both inner race and rolling element fault recognition cases. Recognition of the fault types over other signals (e.g., current or acoustic) for continuous monitoring applications will be part of our future work.

## SUPPLEMENTARY MATERIAL

Bearing faults are mechanical defects and they cause vibration at fault related frequencies. The bearing defect fundamental frequencies can be determined if both bearing geometry and shaft speed are available. These bearings comprise of a set of balls or rolling elements rotating inside an inner and outer ring. In this study, bearing fault detection from raw vibration data is considered. Typical ball bearing geometry is depicted in Figure 4.

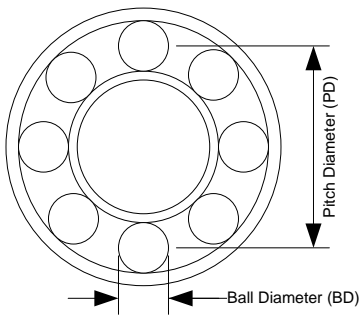

Figure 4: Ball bearing geometry

Bearing faults often cause vibration at the frequencies associated with the fault. These specific frequencies can be used for identifying the different types of faults [27]:

When the motor is running, the cage turns at a linear velocity, which is the mean of the inner and outer race linear velocities. The cage linear velocity can be used to compute the fundamental cage defect, $f_{CD}$:

$$f_{CD} = \frac{V_c}{r_c} = \frac{V_i + V_o}{D_c} \quad (12)$$

where vc, rc, vi, vo, and Dc velocity of the cage, radius of the cage, velocity of the inner race, velocity of the outer race, and diameter of the cage respectively. Then,

$$f_{CD} = \frac{f_i r_i + f_o r_o}{PD} = \frac{1}{PD}\left(f_i \frac{PD - BD\cos\theta}{2} + f_o \frac{PD + BD\cos\theta}{2}\right) \quad (13)$$

where $f_i$, $f_o$, $r_i$, $r_o$, $PD$, $BD$, and $\theta$ are frequency of the inner race, frequency of the outer race, radius of the inner race, radius of the outer race, pitch diameter, ball diameter and contact angle of the ball respectively. For motors, the housing is stationary and the outer race is mounted to the housing. The inner race and the shaft are installed together and both rotate at the same angular speed. As a result, it can be assumed the following:

$$f_o = 0 \text{ and } f_i = f_{rm} \quad (14)$$

where $f_{rm}$ is the mechanical rotor speed in Hertz. Now, equation 2 can be written as

$$f_{CD} = \frac{1}{2} f_{rm}\left(1 - \frac{BD}{PD}\cos\theta\right) \quad (15)$$

The inner race defect frequency, $f_{ID}$, is associated with the rate at which bearing balls traverse a defect point on the inner race. Each ball passes the defect point at the frequency difference of cage and inner race. Since there are *n* balls, the frequency is linearly proportional to the number of balls. The fundamental frequency of the inner race defect, then, can be computed as

$$f_{ID} = n|f_{CD} - f_i| \quad (16)$$

Expanding (16),

$$f_{ID} = n\left|\frac{f_i r_i + f_o r_o}{PD} - f_i\right|$$
$$= n\left|\frac{f_i\left(r_c - \frac{BD\cos\theta}{2}\right) + f_o\left(r_c + \frac{BD\cos\theta}{2}\right)}{PD} - f_i\right| \quad (17)$$

$$= \frac{n}{2}\left|(f_i - f_o)\left(1 + \frac{BD\cos\theta}{2}\right)\right|$$

Substituting $f_0 = 0$ and $f_i = f_{rm}$, $f_{ID}$ becomes

$$f_{ID} = \frac{n}{2} f_{rm}\left(1 + \frac{BD}{PD}\cos\theta\right) \quad (18)$$

The fundamental ball defect frequency, $f_{BD}$, results from the rotation of the ball about its own axis through its center.

The ball defect frequency can be formulated as

$$f_{BD} = \left|(f_i - f_{CD})\frac{r_i}{r_b}\right| = \left|(f_o - f_{CD})\frac{r_o}{r_b}\right| \quad (19)$$
$$= \frac{PD}{2BD}\left|(f_i - f_o)\left(1 - \frac{BD^2 \cos^2 \theta}{PD^2}\right)\right|$$

Substituting $f_0 = 0\ and\ f_i = f_{rm}$, $f_{BD}$ becomes

$$f_{BD} = \frac{PD}{2BD} f_{rm}\left(1 - \frac{BD^2 \cos^2 \theta}{PD^2}\right) \quad (20)$$

Motor fault related frequency components usually show up in close neighborhood of fundamental frequency in motor vibration signal spectrum. Their magnitudes are very small compared to the magnitude of power system fundamental frequency. Therefore, the presence of electrical noise and dominant power system fundamental component in the vibration frequency spectrum complicate the motor fault detection process. Usually notch filters are used for pre-processing of motor current data to suppress power system fundamental frequency in the spectrum. From the analysis of vibration signals, starting from the early level faults more chaotic (noise-like) vibrations with higher amplitudes can be clearly seen as the level of the fault increases. This behavior can also be clearly observed in the corresponding amplitude spectrum plots in Figure 5.

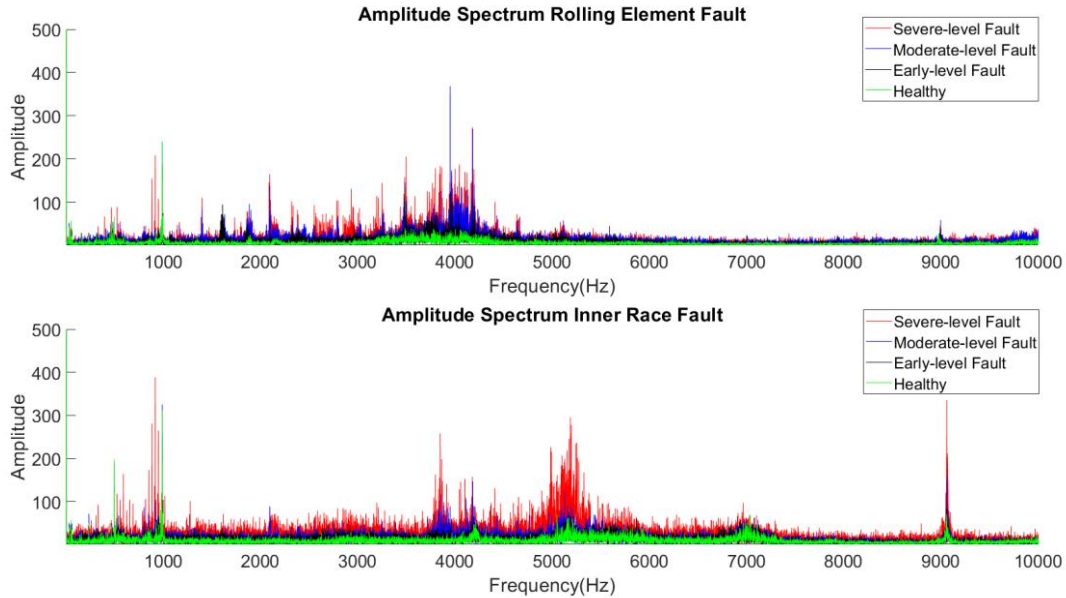

Figure 5: Spectrum of different severity levels of inner race and rolling element faults